\documentclass[sigconf]{acmart}

\usepackage{multirow}
\usepackage{threeparttable}
\usepackage{makecell}
\usepackage{pifont}
\usepackage{graphicx}
\usepackage{diagbox}
\usepackage{booktabs}

\AtBeginDocument{%
  }

\setcopyright{acmlicensed}
\copyrightyear{2018}
\acmYear{2018}
\acmDOI{XXXXXXX.XXXXXXX}
\acmConference[Conference'25]{}{ACM Conference}{2025}
\acmISBN{978-1-4503-XXXX-X/YY/MM}

\begin{document}

\title{Text-Promptable Propagation for Referring Medical Image Sequence Segmentation}

\author{Runtian Yuan$^{1}$, 
        Mohan Chen$^{1}$,
        Jilan Xu$^{1}$,
        Ling Zhou$^{1}$,  
        Qingqiu Li$^{1}$,
}
\author{        
        Yuejie Zhang$^{1,*}$,
        Rui Feng$^{1,*}$, 
        Tao Zhang$^2$,
        Shang Gao$^3$
}
\affiliation{
$^{1}$Fudan University $^{2}$Shanghai University of Finance and Economics $^{3}$Deakin University
\country{Australia}
}

\renewcommand{\shortauthors}{Runtian Yuan et al.}

\begin{abstract}

Referring Medical Image Sequence Segmentation (Ref-MISS) is a novel and challenging task that aims to segment anatomical structures in medical image sequences (\emph{e.g.} endoscopy, ultrasound, CT, and MRI) based on natural language descriptions. 
This task holds significant clinical potential and offers a user-friendly advancement in medical imaging interpretation.
Existing 2D and 3D segmentation models struggle to explicitly track objects of interest across medical image sequences, and lack support for interactive, text-driven guidance. 
To address these limitations, we propose Text-Promptable Propagation (TPP), a model designed for referring medical image sequence segmentation. TPP captures the intrinsic relationships among sequential images along with their associated textual descriptions. 
Specifically, it enables the recognition of referred objects through cross-modal referring interaction, and maintains continuous tracking across the sequence via Transformer-based triple propagation, using text embeddings as queries.
To support this task, we curate a large-scale benchmark, Ref-MISS-Bench, which covers 4 imaging modalities and 20 different organs and lesions.
Experimental results on this benchmark  demonstrate that TPP consistently outperforms state-of-the-art methods in both medical segmentation and referring video object segmentation.
\end{abstract}




\maketitle

\section{Introduction}


\begin{figure*}[ht]
    \centering
    \includegraphics[width=1\linewidth]{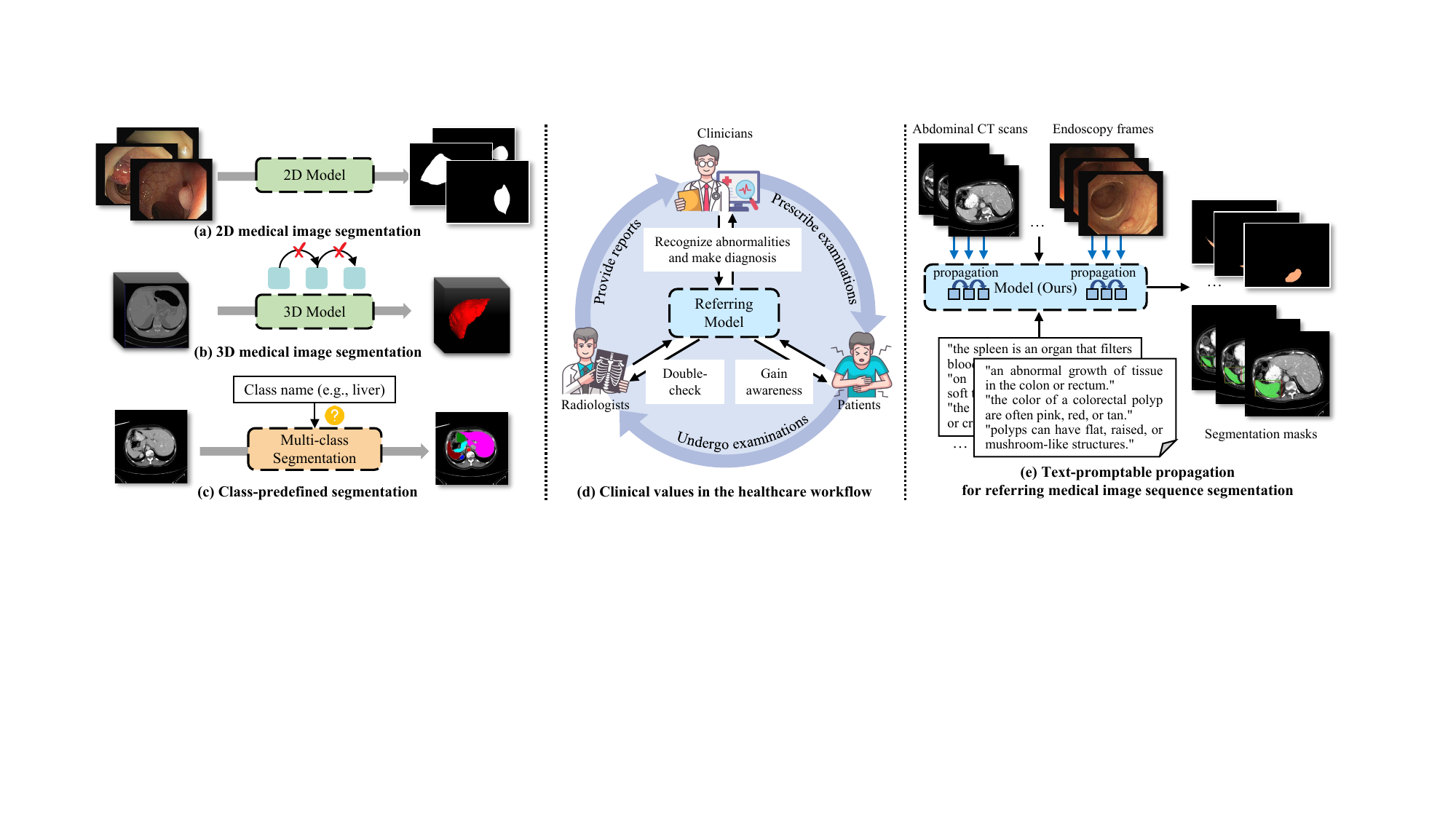}
    \caption{Limitations and motivations. (a) Conventional 2D models do not incorporate temporal context and fail to utilize intrinsic consistencies in medical image sequences. (b) 3D models lack slice-level object representations for modeling continuity. (c) Multi-class segmentation models are limited to predefined classes and cannot use language to specify a particular class. (d) To address these limitations, Referring Medical Image Sequence Segmentation is introduced, offering substantial clinical values. (e) Our TPP leverages medical text prompts to segment referred objects across medical image sequences in both 2D and 3D data.}
    \label{fig:intro}
\end{figure*}

Medical image segmentation plays an important role in modern healthcare by enabling precise delineation of anatomical regions and pathological areas, which is essential for diagnosis, treatment planning, and disease monitoring~\cite{chen2019ldct, cao2023large}. 
Accurate segmentation facilitates quantitative analysis of medical images, supporting early detection of tumors and assessment of organ functionality.

This paper considers medical image sequence segmentation (MISS) task, which involves segmenting medical images from 2D video-based examinations (\emph{e.g.}, endoscopy and ultrasound) and 3D imaging techniques (\emph{e.g.}, CT and MRI). These modalities produce medical image sequences, \emph{i.e.}, temporally or spatially ordered frames or slices that capture the same anatomical structures, including organs and lesions.
Importantly, such sequences are not merely collections of isolated snapshots; rather, they are intrinsically linked, with each frame or slice providing a unique view of the same object from different angles or planes. 
The consistencies among these sequential images are crucial for comprehensive medical analysis and diagnosis.
Modern deep learning models~\cite{ronneberger2015unet,isensee2021nnunet,chen2021transunet,cao2022swinunet,ma2024medsam} have revolutionized image segmentation, however, their capabilities in handling medical image sequences still worth exploration. 

As shown in Figure~\ref{fig:intro} (a)-(c), the main limitations that restrict their real-world clinical utility are three-fold: 
\textbf{First}, most 2D image segmentation models ~\citep{ronneberger2015unet, chen2021transunet} treat frames from video-based examinations or slices from 3D volumes as independent samples, ignoring the inherent spatial and temporal consistencies. 
\textbf{Second}, although existing 3D models~\citep{milletari2016vnet,zhao2023one} can capture correlations between slices, the employed 3D convolutions or attention operations over full 3D patches are computationally expensive and lack the modeling and tracking of objects across sequences.
\textbf{Third}, existing models segment all predefined categories in an image without the ability to incorporate human interaction, limiting their practical value in scenarios where clinicians only care about certain objects.


To address these challenges, we go beyond MISS tasks and instead focus on the more challenging \textbf{Referring Medical Image Sequence Segmentation (Ref-MISS)} task, which requires the model to identify and segment anatomical structures corresponding to given natural language within medical image sequences. 
Enabling users to interact with models 
and specify target structures through language offers several practical benefits, as shown in Figure~\ref{fig:intro} (d): 
(1) radiologists benefit from AI-assisted, text-promptable segmentation results to validate their findings;
(2) clinicians with limited imaging expertise receive clearer explainable visual outputs of lesions from the referring model for decision-making and comprehensive diagnosis;
(3) patients gain from simplified, text-driven visualizations that improve their understanding of medical conditions. 
Ultimately, text-promptable segmentation bridges the gap between visual data and human interpretability, fostering more efficient, accurate, and collaborative healthcare workflows.

To solve Ref-MISS, we propose a novel Text-Promptable Propagation (TPP) model, designed to leverage the intrinsic relationships among sequential images along with their associated textual descriptions, as shown in Figure~\ref{fig:intro} (e). 
TPP integrates two key components: 
(1) \textbf{Cross-modal Referring Interaction.} This component incorporates medical text prompts with vision-language alignment and fusion to recognize referred objects.
Medical text prompts provide critical context by highlighting specific regions of interest and guiding attention. We propose cross-modal referring interaction to integrate prompts, linking medical image sequences with text prompts across vision and language modalities. 
(2) \textbf{Transformer-based Triple Propagation}.
To uniformly model the temporal relationships between 2D frames and cross-slice interactions in 3D volumes, we employ a Transformer-based encoder-decoder architecture, leveraging propagation strategies to track referred objects.

To support this task, we curate a large dataset, \textbf{Ref-MISS-Bench}, from existing public medical datasets, and use Large Language Models (LLMs) to automatically generate text prompts based on different attributes of anatomical structures. The prompts are then validated by senior radiologists. Ref-MISS-Bench is sourced from 18 diverse medical datasets across 4 imaging modalities, including MRI, CT, ultrasound, and endoscopy. It covers 20 different organs and lesions from various regions of the body, and is utilized in both the training and testing stages, as illustrated in Figure~\ref{fig:datasets}.

To summarize, our contributions are as follows:
\begin{itemize}
    \item We focus on the novel task, \textbf{Referring Medical Image Sequence Segmentation (Ref-MISS)}, and establish a strong model, \textbf{T}ext-\textbf{P}romptable \textbf{P}ropagation (TPP), which utilizes medical text prompts to identify referred objects and propagate vision-language information for continuous tracking through sequential images.
    \item We introduce a large-scale benchmark, \textbf{Ref-MISS-Bench}, which covers 4 imaging modalities and 20 anatomical structures. Ref-MISS-Bench consists of 125,487 images from 3,644 sequences in the training set and 41,078 images from 1,061 sequences in the test set, providing a comprehensive data foundation for Ref-MISS task.
    \item Experiments demonstrate that our approach outperforms state-of-the-art methods in 2D/3D/text-guided medical image segmentation and referring video object segmentation, while also incorporating human-interaction capabilities. 
\end{itemize}

\begin{figure*}[]
    \centering
    \includegraphics[width=1\linewidth]{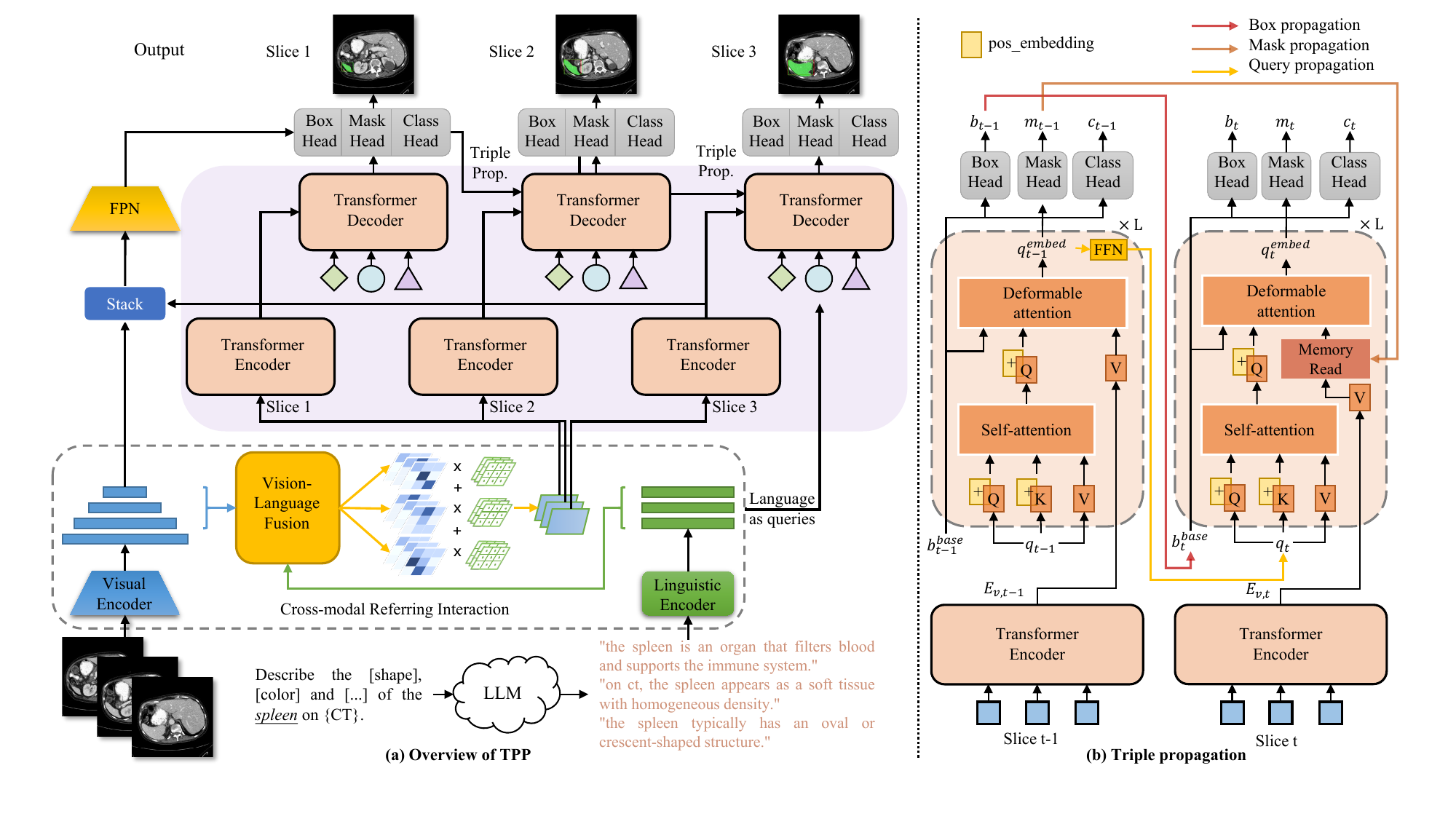}
    \caption{Architecture of our Text-Promptable Propagation for referring medical image sequence segmentation. (a) Overview of TPP. Triple Prop. is short for Triple Propagation. (b) Illustration of Triple Propagation in Transformer decoder, consisting of box-level, mask-level, and query-level propagation. 
    }
    \label{fig:arch}
\end{figure*}

\section{Related work}
\label{RL}

\subsection{Medical Image Segmentation} 
As mentioned earlier, researchers typically apply 2D models~\citep{ronneberger2015unet} for planar images or slices, and 3D models~\citep{cciccek20163dunet, milletari2016vnet} to learn volumetric features implicitly.
\citet{isensee2021nnunet} introduced a versatile, self-adaptive deep learning framework specifically designed for medical image segmentation tasks, extending the U-Net architecture and its 3D version. 
\citet{chen2021transunet} pioneered the combination of Transformer-based architecture with Convolutional Neural Networks (CNNs) for medical image segmentation, applying a slice-by-slice inference on 3D volumes without considering interrelationships among slices. 
Some works~\citep{ji2021pns, painchaud2022echocardiography, lin2023shifting} utilize spatial-temporal cues and \citep{li2023lvit,zhong2023ariadne,bui2024mmiunet} introduce report texts as guidance to enhance segmentation performance. However, these models are limited to specific image modalities and tasks.  

\subsection{Medical Vision-Language Models} Medical vision-language models have achieved success across multiple downstream tasks, including diagnosis classification~\citep{moon2022medvill, medclip, chexzero, lu2023mizero}, lesion detection~\citep{qin2023mvlm, huang2024adapting}, image segmentation~\citep{zhao2023one, li2023lvit}, report generation~\citep{yan2022clinical-bert, BioViL-T}, and visual question answering~\citep{singhal2023med-palm, moor2023med-flamingo}. 
\citet{qin2023mvlm} designed auto-generation strategies for medical prompts and transferred large vision language models for medical lesion detection. 
\citet{liu2023clip-driven} incorporated text embedding learned from Contrastive Language-Image Pre-training (CLIP) to segmentation models.
\citet{zhao2024biomedparse} proposed BiomedParse, a biomedical foundation model that can jointly conduct segmentation, detection and recognition across nine imaging modalities.
\citet{zhao2023one} built a model based on Segment Anything Model~\citep{kirillov2023sam} for medical scenarios driven by text prompts, but the model focused on 3D medical volume segmentation and did not consider the sequential relationships between scans. 
To the best of our knowledge, we are the first to use medical text prompts to specify segmentation targets across medical image sequences.


\subsection{Referring Video Object Segmentation}
\citet{gavrilyuk2018actor} were the first to propose inferring segmentation from a natural language input, extending two popular actor and action datasets with natural language descriptions.
\citet{seo2020urvos} constructed the first large-scale referring video object segmentation (RVOS) dataset and proposed a unified referring video object segmentation network.
\citet{wu2022referformer} and \citet{botach2022mttr} presented Transformer-based RVOS frameworks, enabling end-to-end segmentation of the referred object.
\citet{wu2023onlinerefer} designed explicit query propagation for an online model. 
\citet{luo2024soc} aggregated inter- and intra-frame information via a semantic integrated module and introduced a visual-linguistic contrastive loss to apply semantic supervision on video-level object representations.
\citet{yan2024referred} enabled multi-modal references to capture multi-scale visual cues and designed inter-frame feature communication for different object embeddings for tracking along the video.

Inspired by these works, the Referring Medical Image Sequence Segmentation task processes both 2D and 3D medical data into image sequences, enabling in-depth exploration of sequence-level consistency guided by text prompts.

\section{Methodology}


\subsection{Problem Formulation}
This paper tackles the Referring Medical Image Sequence Segmentation (Ref-MISS) task. Formally, given $T$ frames or slices $\{I_t \in \mathbb{R}^{{3}\times{H}\times{W}}\}_{t=1}^T$ from a medical image sequence 
and $N_p$ medical text prompts $\{P_i\}_{i=1}^{N_p}$ (Section~\ref{benchmark}), the referring model $\mathcal{M}$ aims to predict the segmentation masks $\{\hat{m_t} \in\{0,1\}^{{H}\times{W}}\}_{t=1}^T$ for the referred object corresponding to the prompts, which can be formulated as: 
\begin{equation}
    \{\hat{m_t}\}_{t=1}^T = \mathcal{M}\left(\{I_t\}_{t=1}^T, \{P_i\}_{i=1}^{N_p}\right).
\end{equation}
An overview of our framework is illustrated in Figure~\ref{fig:arch} (a). The referring model $\mathcal{M}$ comprises two core components: \textbf{Cross-Modal Referring Interaction} (Section~\ref{method1}) to recognize the referred objects, and \textbf{Transformer-based Triple Propagation} (Section~\ref{method2}) to maintain continuous tracking across sequences. The training and inference procedures are described in Section~\ref{method3}.

\subsection{Cross-Modal Referring Interaction}
\label{method1}
\paragraph{Visual Feature Extraction.}
The visual encoder $\phi_v$ takes the medical image sequence $\{I_t\}_{t=1}^T$ as input, and encodes them in a per-frame manner. The visual encoder outputs multi-scale features $F_v$ for each image, which is a set of feature maps:
\begin{equation}
    \{f_{v}^{l}\}_{l=1}^{4}=\phi_v(I_t) \in\mathbb{R}^{{C^{l}}\times {H^{l}} \times{W^{l}}},
\end{equation}
where $C^l$, $H^l$ and $W^l$ denote the channel dimension, height, and width of the feature map at the $l^{th}$ level, respectively. 

\paragraph{Textual Feature Extraction.}
The linguistic encoder $\phi_t$ takes the medical text prompts $\{P_i\}_{i=1}^{N_p}$ as input, encodes each prompt independently, and outputs the textual feature $F_p$, which is a set of word-level embeddings $\{f_{p}^{i}\}_{i=1}^{N_p}$. The encoding process of each prompt $P_i$ is defined as: 
\begin{equation}
    f_p^{i} = \phi_t(P_i) \in\mathbb{R}^{{Len_i}\times C},
\end{equation}
where $Len_i$ and $C$ denotes the length of sentence embedding and hidden dimension, respectively.


\paragraph{Vision-Language Alignment and Fusion.}
After obtaining the visual and textual features, we align and fuse them to enhance the model's focus on the referred objects 
and identify the most relevant prompt for each image clip. 
This process involves three key steps. 

(1) \textbf{Cross-modal attention.} For each image, we apply Multi-Head Attention (MHA) mechanisms between the visual feature maps at the last three levels ($l=\{2,3,4\}$) and the word-level embeddings from the text prompts. This produces a set of proposal features:
\begin{equation}
    A_{}^{l,i} = \operatorname{MHA} \left(f_{v}^{l}, f_{p}^{i} \right), 
\end{equation}    
where $A_{}^{l,i}$ represents the attention output between the $l$-th visual feature map and the $i$-th text prompt. Each prompt (\emph{i.e.}, $P_1, P_2, P_3$)
yields its own set of proposals, denoted as $\displaystyle \mathbb{A}$, $\displaystyle \mathbb{B}$, $\displaystyle \mathbb{C}$, respectively. This enables modeling of complex vision-language dependencies. 

(2) \textbf{Weighted fusion of proposals.} To identify the referred object, i.e. $\displaystyle \mathbb{A} \cap \displaystyle \mathbb{B} \cap \displaystyle \mathbb{C}$, we flatten each proposal and apply a three-layer Multi-Layer Perceptron (MLP) to compute prompt-specific relevance weights:
\begin{equation}
    W_{}^{l,i} = \operatorname{Softmax} \left(\operatorname{MLP} \left(A^{l,i}\right)\right),
\end{equation}
which are then used to perform a weighted sum over prompts:
\begin{equation}
    F'_{v} = \left\{\sum_{i=1}^{N_p} f_{v}^{l} \cdot A^{l,i} \cdot W^{l,i}\right\}_{l=2}^{4}.
\end{equation}
This step generates the fused visual features, integrating the most pertinent aspects of text prompts with the visual data.

(3) \textbf{Prompt selection for query input.} For textual features, we select the most relevant prompt with the highest weight score produced by the feature maps at the first level ($l=\{1\}$). The selected prompt feature $F'_{p}$ is then used as the query input to the Transformer decoder.
\begin{equation}
    \hat{w} = \mathop{\arg\max}\limits_{i\in\{1,\dots,N_p\}}\left(W^{l=1,i}\right),
    F'_{p}  = f_p^{\hat{w}}.
\end{equation}


\begin{figure*}
    \centering
    \includegraphics[width=1\linewidth]{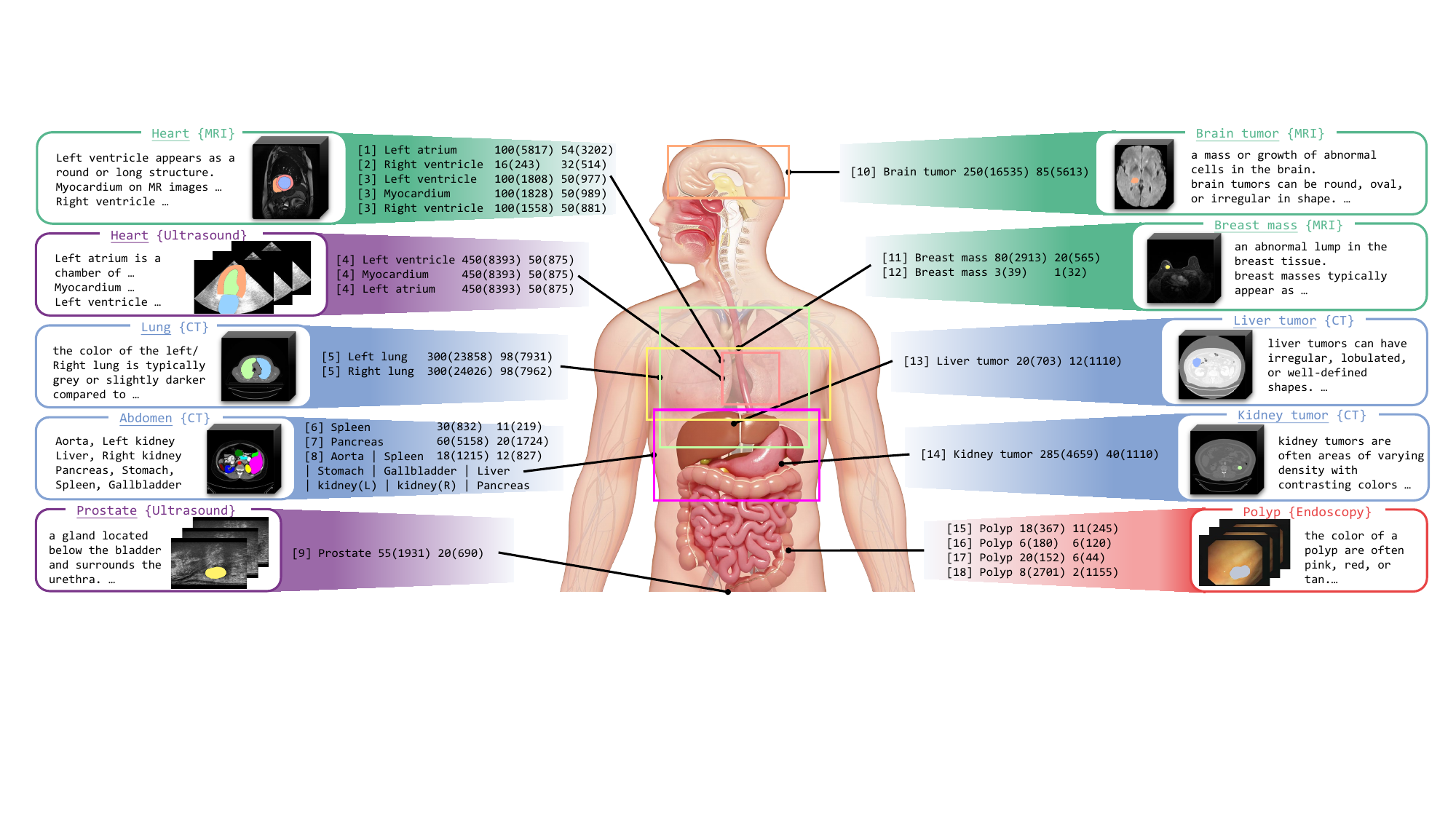}
    \caption{An illustration of focus areas in Ref-MISS-Bench. Each colored block represents specific organ/lesion class from corresponding [dataset], along with number of training and testing cases (images).}
    \label{fig:datasets}
\end{figure*}

\subsection{Transformer-based Triple Propagation}
\label{method2}

\paragraph{Transformer.} Our Transformer architecture is adapted from Deformable DETR~\citep{zhu2021deformable}. For each image $I_t$, the Transformer encoder takes the flattened visual features $F'_{v,t}$ and 2D positional encoding as input, producing encoded output $E_{v,t}$ through multi-scale deformable attention and several feed-forward layers. The output of the Transformer encoder $E_{v,t}$ and the textual feature of the selected prompt $F'_{p,t}$ are then fed into the Transformer decoder. We repeat $F'_{p,t}$ $N_q$ times to introduce $N_q$ queries, denoted as $q_t$. Meanwhile, each image receives sequential cues from the previous frame (except for the first image) in temporal order. The Transformer decoder thus generates $N_q$ embeddings for each image, denoted as $q_t^{embed}$. 

\paragraph{Prediction Heads.} Three prediction heads are constructed following the Transformer decoder. The output embeddings from the Transformer decoder, $q_t^{embed}$, are then processed by these prediction heads. 
(1) The \textbf{box head} consists of a three-layer feed-forward network (FFN) with ReLU activation, except for the last layer, which predicts the box offset. This offset is added to the base box coordinates to determine the location of the referred object, denoted as $b_t$.
(2) The \textbf{mask head} is implemented by dynamic convolution~\citep{tian2020conditional}. It takes multi-scale features from the feature pyramid network (FPN) $f_m$, concatenates them with relative coordinates, and uses a controller to generate convolutional parameters $\theta_t$. Conditional convolution is then applied to the visual features to generate $N_q$ segmentation masks $m_t$.
\begin{equation}
    \theta_t = \operatorname{Controller} \left(q_t^{embed}\right),
\end{equation}
\begin{equation}
    \left\{m_t^{i}\right\}_{i=1}^{N_q} = \left\{\phi^{i} \left(f_{m};\theta_t^{i}\right)\right\}_{i=1}^{N_q}.
\end{equation}
Here, the controller is also a three-layer FFN with ReLU activation. $\phi^i$ represents three $1\times1$ convolutional layers with 8 channels per query, using parameters $\theta_t^i$ generated by the controller. 
(3) Since our text prompts contain class information, the \textbf{class head} indicates whether the object is referred by the text prompt. 

\paragraph{Triple Propagation.} 
Medical image sequences often exhibit high temporal consistency in appearance and spatial structure. To exploit this, we propagate the box, mask, and query embeddings derived from the previous image to inform predictions for the current image, as depicted in Figure~\ref{fig:arch} (b). This triple propagation 
enhances  robustness and accuracy in medical image sequence analysis. 

Given previous predictions $y_{t-1} = \{b_{t-1}^i, m_{t-1}^i, c_{t-1}^i\}_{i=1}^{N_q}$, 
we choose the best prediction $\{b_{t-1}^{\hat{n}}, m_{t-1}^{\hat{n}}, c_{t-1}^{\hat{n}}\}$, which is of the highest class score. Consequently, except for the first image, which has $N_q$ queries, subsequent images only receive one query propagated from the previous best prediction. 

{\textbf{Box-level} Propagation.} The box coordinates from the previous image $b_{t-1}^{\hat{n}}$ provide a valuable reference for estimating the location of the referred object in the current image. We use these coordinates as the initial box for the current image, \emph{i.e.} $b_{t}^{base}$, leveraging the spatial continuity to provide a strong prior for localization. Box-level propagation improves precision by refining the search around a plausible region.

{\textbf{Mask-level} Propagation.} Similarly, the visual features encoded by the Transformer encoder $E_{v,t-1}$ and the segmentation mask $m_{t-1}^{\hat{n}}$ from the previous image offer valuable semantic context that can aid in analyzing the current image. To effectively utilize this prior knowledge, we employ a memory-read mechanism that generates key and value maps for the memory. The memory map $M_{t-1}$ is a concatenation of $m_{t-1}^{\hat{n}}$ and the first-level of $E_{v-1,t}$, and the memory read operation is defined as:
\begin{equation}
    M_{t-1} = \operatorname{Concat} \left(m_{t-1}^{\hat{n}}, E_{v,t-1}^{l=2}\right),
\end{equation}
\begin{equation}
    K = \psi \left( M_{t-1} \right), V = \varphi \left( M_{t-1} \right),
\end{equation}
\begin{equation}
    E_{v,t}^{l=2} = \operatorname{Softmax} \left(\frac{E_{v,t}^{l=2} K}{\sqrt{C^{l=2}}}\right)V,
\end{equation}
where $\psi$ and $\varphi$ are two parallel $3\times3$ convolutional layers. The first level of $E_{v,t}$ is now a memory-read map. It is concatenated with feature maps of other levels and then fed into the deformable attention module in the Transformer decoder after flattening.

{\textbf{Query-level} Propagation.} Having confirmed the 
query index $\hat{n}$, we propagate the corresponding output query embedding $q_{t-1}^{embed}$ to the current image. Here, we use a three-layer FFN to transform the embedding to $q_{t}$. This query-level propagation allows for the transmission of embedded context for the same target.

\subsection{Training and Inference}
\label{method3}

\paragraph{Training.} We have $N_q$ predictions $y_t = \{b_t^i, m_t^i, c_t^i\}_{i=1}^{N_q}$ for each image, where $b_t^i \in \mathbb{R}^4$, $m_t^i \in \mathbb{R}^{\frac{H}{4} \times \frac{W}{4}}$, and $c_t^i \in \mathbb{R}^1$ represent the predicted box location, segmentation mask, and probability of the referred object, respectively. The ground-truth, in the same format, is denoted as $Y_t = \{B_t, M_t, C_t\}$. We compute a matching loss $\mathcal{L}_{match}$ to find the best prediction:
\begin{equation}
\begin{aligned}
\mathcal{L}_{match,t}\left(y_t, Y_t\right) & = \lambda_{box}\mathcal{L}_{box}\left(y_t, Y_t\right) \\
& + \lambda_{mask}\mathcal{L}_{mask}\left(y_t, Y_t\right) \\
& +  \lambda_{cls}\mathcal{L}_{cls}\left(y_t, Y_t\right),
\end{aligned}
\end{equation}
\begin{equation}
    \hat{n}_{q,t} = \mathop{\arg\min}\limits_{i\in\{1,\dots,N_q\}}\left(\mathcal{L}_{match,t}\right),
\end{equation}
where $\lambda_{box}$, $\lambda_{mask}$, and $\lambda_{cls}$ are loss coefficients. $\mathcal{L}_{box}$ is implemented as the sum of L1 loss and GIoU loss, $\mathcal{L}_{mask}$ combines Dice loss and binary mask focal loss, and $\mathcal{L}_{cls}$ is focal loss. $\hat{n}_{q,t}$ represents the query index of the best prediction.
The network is optimized by minimizing the sum of $\mathcal{L}_{match,t}$ for the best predictions across all $T$ images.
\begin{equation}
    \mathcal{L} = \frac{1}{T} \sum_{t=1}^{T} \mathcal{L}_{match,t}^{\hat{n}_{q,t}}.
\end{equation}

\paragraph{Inference.} During inference, we select the query with the highest class score as the best prediction, which can be formulated as:
\begin{equation}
    \hat{n'}_{q,t} = \mathop{\arg\max}\limits_{i\in\{1,\dots,N_q\}}\left(c_t^i\right).
\end{equation}
The final segmentation masks for each image $\{\hat{m_t}\}_{t=1}^T$ are selected using the query index $\hat{n'}_{q,t}$ from the $N_q$ predictions $\{m_t^i\}_{i=1}^{N_q}$. 
Due to our propagation strategy, the best prediction of the first image is propagated to subsequent images, leading to a single query for each of the remaining images. Therefore, for $t > 1$, the final mask simplifies to $\hat{m_t} = m_t$. 

\begin{table*}[]
\caption{Comparison with task-specific medical image segmentation methods. Numbers in \textbf{bold} indicate the best and \underline{underlined} ones represent the second best. ${^1}$Average of ACDC and CAMUS, ${^2}$Average of BTCV, Pancreas-CT, and Spleen segmentation dataset. $^{3}$Average of Breast Cancer DCE-MRI
Data and RIDER. $^{4}$Average of CVC-ClinicDB, CVC-ColonDB, ETIS, and ASU-Mayo.}
\label{comparison_medical}
\begin{center}
\renewcommand\arraystretch{1.1} 
\scalebox{1}{
\begin{tabular}{cccccccccccc}
\toprule
{\bf Method} & \textbf{Type} & {Heart${^1}$} &  {Lung\tnote{}} & \makecell[c]{Abd-\\omen${^2}$} &  \makecell[c]{Pro-\\state} &  {\makecell[c]{Brain \\tumor}} &  \makecell[c]{Breast \\mass${^3}$} &  \makecell[c]{Liver \\tumor} &  \makecell[c]{Kidney \\tumor}  &  {Polyp${^4}$} &  \textbf{ Overall}\\ 
\cmidrule(r){1-2} \cmidrule(){3-11} \cmidrule(l){12-12}
UNetR~\citep{hatamizadeh2022unetr} & Image-only & - & 84.69 & 70.33 & - & 76.15& 61.23 & 63.42 & 74.21 & - & 71.67 \\
Swin-UNet~\citep{cao2022swinunet} & Image-only & - & 85.40 & 70.96 & - & 75.48 & 60.27 & 64.90 & 74.38 & - & 71.90 \\
nn-UNet~\citep{isensee2021nnunet} & Image-only & 85.63 & 81.59 & 72.31 & 89.73 & 76.57 & 56.80 & 74.89 & 77.06 & 47.99 & 73.62 \\
MedSAM~\citep{ma2024medsam} & Image-only & 85.98 & 86.57 & 73.94 & 89.91 & 77.98 & 62.34 & 62.91 & 77.47 & 75.50 & 76.96  \\
\midrule
LViT~\citep{li2023lvit} & Text-image  & 79.58 & 83.87 & 60.45 & 90.22 & 75.67 & 48.87 & 63.99 & 64.77 & 58.63 & 69.56 \\
LGMS~\citep{zhong2023ariadne} & Text-image  & 83.58 & 86.08 & 70.20 & 91.61 & 78.06 &51.80 & 64.03 & 74.48 & 61.94 & 74.64 \\
MMI~\citep{bui2024mmiunet} & Text-image  & 82.60 & 85.54 & 64.96 & 90.24 & 76.71 & 61.77 & 64.96 & \textbf{78.10} & 71.30 & 75.13 \\
\midrule
 Ours & Text-image & \textbf{87.19} & \textbf{88.77}  & \textbf{72.80} & \textbf{93.13} & \textbf{78.24} & \textbf{65.40} & \textbf{65.27}  & \underline{77.73}  & \textbf{75.56} & \textbf{78.23}\\
\bottomrule
\end{tabular}
}
\end{center}
\end{table*}

\begin{table*}[]
\caption{Comparison with state-of-the-art methods on referring video object segmentation.}
\label{comparison_rvos}
\begin{center}
\renewcommand\arraystretch{1.1} 
\scalebox{1}{
\begin{tabular}{cccccccccccc}
\toprule
{\bf Method}& {\bf Backbone} &  {Heart${^1}$} &  {Lung\tnote{}} & \makecell[c]{Abd-\\omen${^2}$} &  \makecell[c]{Pro-\\state} &  {\makecell[c]{Brain \\tumor}} &  \makecell[c]{Breast \\mass${^3}$} &  \makecell[c]{Liver \\tumor} &  \makecell[c]{Kidney \\tumor}  &  {Polyp${^4}$} &  \textbf{ Overall}\\ 
\cmidrule(r){1-2} \cmidrule(){3-11} \cmidrule(l){12-12}
URVOS~\citep{seo2020urvos} & ResNet-50 & 83.92  & 84.61  & 60.19 & 91.92 & 74.59 & 55.91  & 27.43  & 72.24 & 66.17 & 68.55 \\
ReferFormer~\citep{wu2022referformer} & ResNet-50 & 86.29  & 84.19 & 72.12 & 89.79 & 76.60   & 60.70  & 47.43  & 61.75  & 62.75 & 71.29 \\
OnlineRefer~\citep{wu2023onlinerefer} & ResNet-50 & 83.93 & 85.27 & 63.48 & 91.69 & 77.55 & 64.81 & 39.70 & 74.75 & 72.77 & 72.66 \\
 Ours & ResNet-50  & \textbf{87.19} & \textbf{88.77}  & \textbf{72.80} & \textbf{93.13} & \textbf{78.24} & \textbf{65.40} & \textbf{65.27}  & \textbf{77.73}  & \textbf{75.56} & \textbf{78.23}\\
\midrule
ReferFormer~\citep{wu2022referformer} & Swin-L & 84.12 & 82.56 & 66.05 & 90.58  & 76.89 & 61.53 & 57.43 & 78.31 & 67.35 & 73.87 \\
OnlineRefer~\citep{wu2023onlinerefer} & Swin-L & 84.37 & 83.59 & 60.39 & 90.72  & 77.46 & 57.22  & 54.50 & 69.91  & \textbf{78.47} & 72.96 \\
 Ours & Swin-L & \textbf{84.47}& \textbf{84.96} & \textbf{66.41} & \textbf{91.54} & \textbf{77.96} & \textbf{65.90} & \textbf{59.32} & \textbf{79.27} & \underline{77.56} & \textbf{76.38} \\
\midrule
SOC~\citep{luo2024soc} & V-Swin-T & 81.76 & 84.84 & 62.55 & 86.42  & 75.55  & \textbf{61.57} & 35.30  & 70.01  & 60.04 & 68.67 \\
MTTR~\citep{botach2022mttr} & V-Swin-T & 84.80 & 84.92  & 64.23  & 89.96  & 76.21 & 57.74  & 53.68  & 67.31  & 71.12 &72.22 \\
  Ours & V-Swin-T & \textbf{84.98} & \textbf{85.19} & \textbf{65.57} & \textbf{92.34}& \textbf{77.37}  & \underline{59.17} & \textbf{54.26} 
 & \textbf{76.07} & \textbf{77.11} & \textbf{74.67} \\
\bottomrule
\end{tabular}
}
\end{center}
\end{table*}

\section{Benchmark Construction}
\label{benchmark}
\paragraph{Dataset Curation.}
Ref-MISS-Bench is curated from 18 medical image sequence datasets with 20 anatomical structures across 4 different imaging modalities, as shown in Figure~\ref{fig:datasets}. 

These datasets are categorized by imaging modalities as follows:
(1) \textbf{MRI datasets}.
2018 Atria Segmentation Data~\citep{xiong2021la},
RVSC~\citep{petitjean2015rvsc}, 
ACDC~\citep{bernard2018acdc},
BraTS 2019~\citep{menze2014brats1, bakas2017brats2, baid2021brats3},
Breast Cancer DCE-MRI Data~\citep{zhang2023breast_mri}, and RIDER~\citep{Meyer2015RIDER}.
(2) \textbf{CT datasets}.
Thoracic cavity segmentation dataset~\citep{Aerts2019NSCLC},
spleen segmentation dataset~\citep{simpson2015spleen},
Pancreas-CT~\citep{roth2015pancreas},
the abdomen part of BTCV~\citep{landman2015btcv},
LiTS~\citep{bilic2023lits}, and 
KiTS 2023~\citep{heller2021kits19, heller2023kits21},
(3) \textbf{Ultrasound datasets}.
CAMUS~\citep{leclerc2019camus} (also known as echocardiography), and
Micro-Ultrasound Prostate Segmentation Dataset~\citep{jiang2024microsegnet}.
(4) \textbf{Endoscopy datasets}. 
CVC-ClinicDB~\citep{bernal2015cvc600},
CVC-ColonDB~\citep{bernal2012cvc300},
ETIS~\citep{silva2014etis}, and
ASU-Mayo~\citep{tajbakhsh2015asu}.
For all datasets, videos are converted into frames and 3D volumes are converted into 2D slices. In total, there are 3,644 sequences (125,487 images) for training and 1,061 sequences (41,078 images) for testing. 

\paragraph{Prompt Acquisition.} 
We adopt large language models to automatically generate medical text prompts. These medical text prompts are then proofread by senior radiologists.
The instruction template is as follows:
``You are a medical expert. Describe the [attribute 1], [attribute 2], ..., and [attribute $N_p$] of the \underline{\textit{anatomical structure}} on \{modality\} in one sentence each.'' 

Using this template, we obtain $N_p$ prompts for the target object (\emph{i.e.}, anatomical structure) that is expected to be segmented. Here, $N_p$ is set to 3, with [attribute 1]=[profile], [attribute 2]=[shape], and [attribute 3]=[color]. The attribute [profile] characterizes organ functions and defines lesions, while attributes [color] and [shape] describe the morphological aspects of the object. Detailed prompts can be found in supplementary materials.

\section{Experiments}

\subsection{Experimental Settings}




We train a universal model on Ref-MISS-Bench and maintain the original training and testing splits, ensuring that each sequence appears in only one split. 
Data augmentation techniques include random horizontal flipping, random resizing, random cropping, and photometric distortion. All images are resized to a maximum length of 640 pixels.
Segmentation performance is evaluated using the Dice score.
The coefficients for the loss terms are set as follows: $\lambda_{L1} = 5$, $\lambda_{giou} = 2$, $\lambda_{dice} = 5$, $ \lambda_{focal} = 2$, and $\lambda_{cls} = 2$. We adopt 4 encoder layers and 4 decoder layers in the Transformer. The initial query number $N_q$ is set to 5. Both the hidden dimension of the Transformer and the channel dimension of text prompts are $C=256$. 
During training, 3 temporal images from a sequence are randomly sampled and fed into the model at each iteration. Our model is trained on 2 RTX 3090 24GB GPUs, with AdamW optimizer and an initial learning rate of $10^{-5}$ for 5 epochs. The learning rate decays by 0.1 at the $3^{rd}$ epoch.

\begin{figure*}[h]
    \centering
    \includegraphics[width=.8\linewidth]{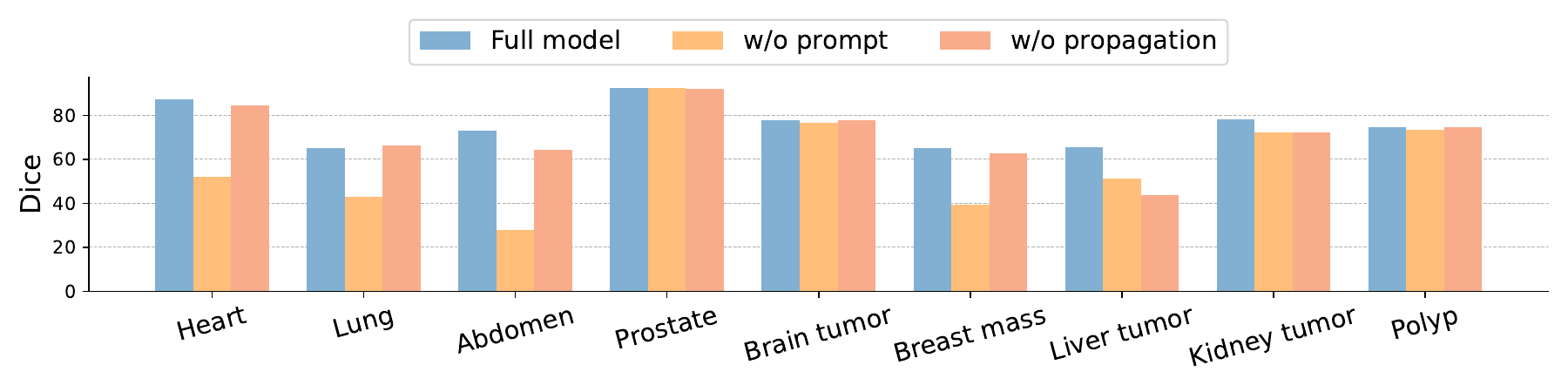}
    \caption{Ablation studies on text prompts and propagation strategies. Dice scores are provided for full model, without prompt, and without propagation, respectively. 
    }
    \label{fig:ab}
\end{figure*}

\subsection{Results}
\subsubsection{Comparison to State-of-the-art in Medical Domain}
To better organize and present the datasets, we categorize the organ datasets into four anatomical groups: heart, lung, abdomen, and prostate. We then compute the average metrics for each group, allowing us to identify strengths and weaknesses across different anatomical regions. Detailed experimental results for each category are provided in supplementary materials. Table~\ref{comparison_medical} shows comparison results with UNetR~\citep{hatamizadeh2022unetr}, Swin-UNet~\citep{cao2022swinunet}, nn-UNet~\citep{isensee2021nnunet}, MedSAM~\citep{ma2024medsam}, LViT~\citep{li2023lvit}, LGMS~\citep{zhong2023ariadne}, and MMI~\citep{bui2024mmiunet}. Among them, UNetR and Swin-UNet are 3D models, while LViT, LGMS, and MMI utilize multi-modal inputs combining images with text annotations. We train and evaluate \textbf{separate models} for these task-specific methods on each anatomical structure. Experimental results demonstrate superior performance of our \textbf{universal model} over them.

\subsubsection{Comparison to State-of-the-art on RVOS} 
\begin{figure*}
    \centering
    \includegraphics[width=1\linewidth]{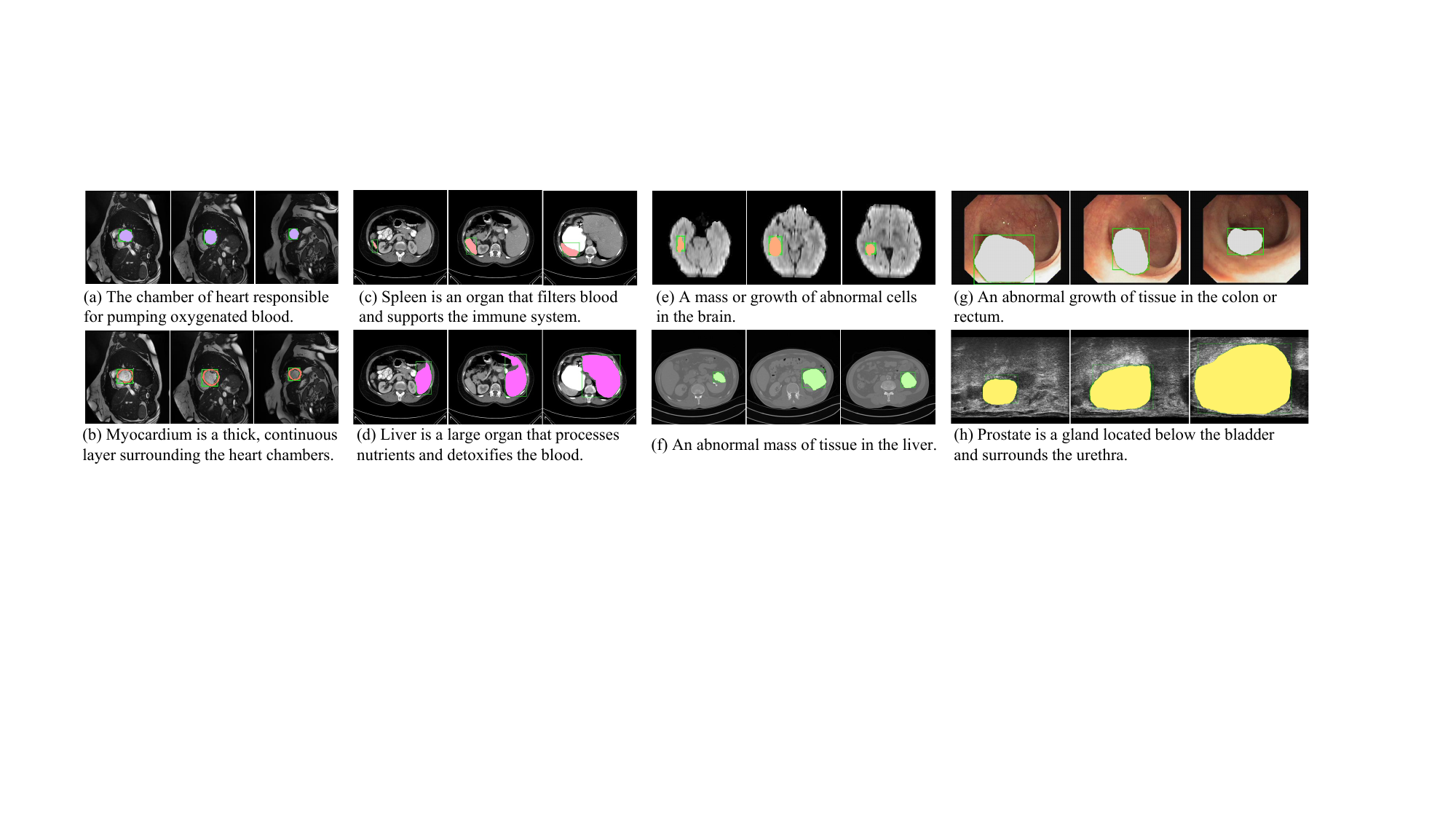}
    \caption{
    Visualization of segmentation results for different structures and modalities.
    (a) and (b) display the results of left atrium and myocardium in the same MRIs, respectively. (c) and (d) show spleen and liver in the same CT slices, respectively. From (e) to (h), 
    visualizations are: brain tumor in MRI, liver tumor in CT, polyp in endoscopy, and prostate in ultrasound.
    }
    \label{fig:visual}
\end{figure*}

We compare our method with state-of-the-art approaches on referring video object segmentation, including URVOS~\citep{seo2020urvos}, ReferFormer~\citep{wu2022referformer}, OnlineRefer~\citep{wu2023onlinerefer}, MTTR~\citep{botach2022mttr}, and SOC~\citep{luo2024soc}. Comparison results for both organs and lesions are shown in Table~\ref{comparison_rvos}.
For feature extraction, we implement multiple visual backbones, including ResNet~\citep{he2016resnet}, Swin Transformer~\citep{liu2021swin}, and Video Swin Transformer~\citep{liu2022video}.
Notably, the performance for organ detection is higher than that for lesion detection. This discrepancy can be attributed to the smaller size and more homogeneous appearance of lesions, which makes them inherently more challenging to identify. 
Our approach consistently outperforms previous methods across all three backbones, especially on lesion datasets. For instance, in segmenting liver and kidney tumors, our model with a ResNet-50 backbone achieves average Dice scores of 65.27\% and 77.73\%, which are 17.84 and 15.98 points higher than the previous state-of-the-art work, ReferFormer.
Visual results of our TPP are shown in Figure~\ref{fig:visual}.

\begin{table}[t]
\caption{Comparison with SAM 2 series.}
\centering
\renewcommand\arraystretch{1.1} 
\scalebox{1}{
\begin{tabular}{ccc}
\toprule
{\textbf{\makecell[c]{Prompter + Segmenter}}} & {\textbf{Organ}} & {\textbf{Lesion}} \\ 
\cmidrule(r){1-1} \cmidrule(){2-3}
{G. DINO + SAM 2} & 12.46 & 10.10 \\
{TPP + SAM 2}  & 53.45 {\scriptsize\textcolor{red}{(+40.99)}} & 54.55 {\scriptsize\textcolor{red}{(+44.45)}} \\
Ours (TPP + TPP)  & 80.77 {\scriptsize\textcolor{red}{(+68.31)}} & 72.69 {\scriptsize\textcolor{red}{(+62.59)}} \\ 
\bottomrule
\end{tabular}}
\label{tab:sam2}
\end{table}

\begin{table}[t]
\caption{Few-shot performance.}
\centering\renewcommand\arraystretch{1.1} 
\scalebox{1}{
\begin{tabular}{cccc}
\toprule
{\textbf{Method}}  & {\textbf{\makecell[c]{Right ventricle}}} & {\textbf{\makecell[c]{Breast mass}}} & {\textbf{Polyp}} \\
\cmidrule(r){1-1} \cmidrule(){2-4}
{Full data} & 81.97 & 61.96 & 82.19 \\
{One-shot} & 75.63 {\scriptsize\textcolor{blue}{(-6.34)}} & 59.88 {\scriptsize\textcolor{blue}{(-2.08)}} &  81.55 {\scriptsize\textcolor{blue}{(-0.64)}} \\
{Zero-shot} &  71.13 {\scriptsize\textcolor{blue}{(-10.84)}}  & 57.18 {\scriptsize\textcolor{blue}{(-4.78)}} &  80.97 {\scriptsize\textcolor{blue}{(-1.22)}} \\
\bottomrule
\end{tabular}}
\label{tab:zero}
\end{table}

\subsubsection{Comparison to SAM 2} The Segment Anything Model 2~\citep{ravi2024sam2}
serves as a foundational model for promptable visual segmentation in images and videos. As it currently lacks support for text prompts, we utilize a community-developed version, Grounded SAM 2~\citep{liu2023dino}, which enables video object tracking with text inputs. This model uses box outputs from Grounding DINO as prompts for SAM 2's video predictor, effectively merging SAM 2's tracking capabilities with Grounding DINO for open-set video object segmentation. 
Despite this integration, it achieves average Dice scores of only 12.46\% for organs and 10.10\% for lesions, indicating its limited understanding of medical text prompts. 

To address this, we utilize the mask predictions of the first image in the sequences generated by our TPP as mask prompts for SAM 2. 
This leads to substantial improvements, with average Dice scores increasing to  53.45\% for organs and 54.55\% for lesions. As shown in Table~\ref{tab:sam2}, our TPP demonstrates superiority over Grounding DINO in text grounding ability, and surpasses SAM 2 in object tracking capabilities due to the triple propagation strategy. 

\subsubsection{Zero-/One-shot Performance} 
To validate the zero-shot performance of our approach on unseen datasets, we exclude RVSC (right ventricle), RIDER (breast mass), and CVC-ColonDB (polyp) from the training datasets and evaluate the trained model on these datasets directly. 
As shown in Table~\ref{tab:zero}, the Dice scores for breast mass and polyp decrease by only 4.78 and 1.22 points, respectively, compared to full-data training.
In the one-shot setting, we add a single sequence from each of the three datasets mentioned above into the training set. The results show that one-shot performance on polyp is comparable to full-data training, highlighting the model's robust generalization ability.

\subsection{Ablation studies}

Cross-modal referring interaction and the propagation strategy are critical components of our approach to referring medical image sequence segmentation. Figure~\ref{fig:ab} illustrates that medical text prompts are particularly essential for accurately identifying organs located in the heart, lungs, and abdomen. Moreover, for extremely small lesions, such as breast masses and liver tumors, our propagation strategy significantly reduces the occurrence of false negatives, resulting in substantial enhancements.

\begin{figure}[]
    \centering
    \includegraphics[width=1\linewidth]{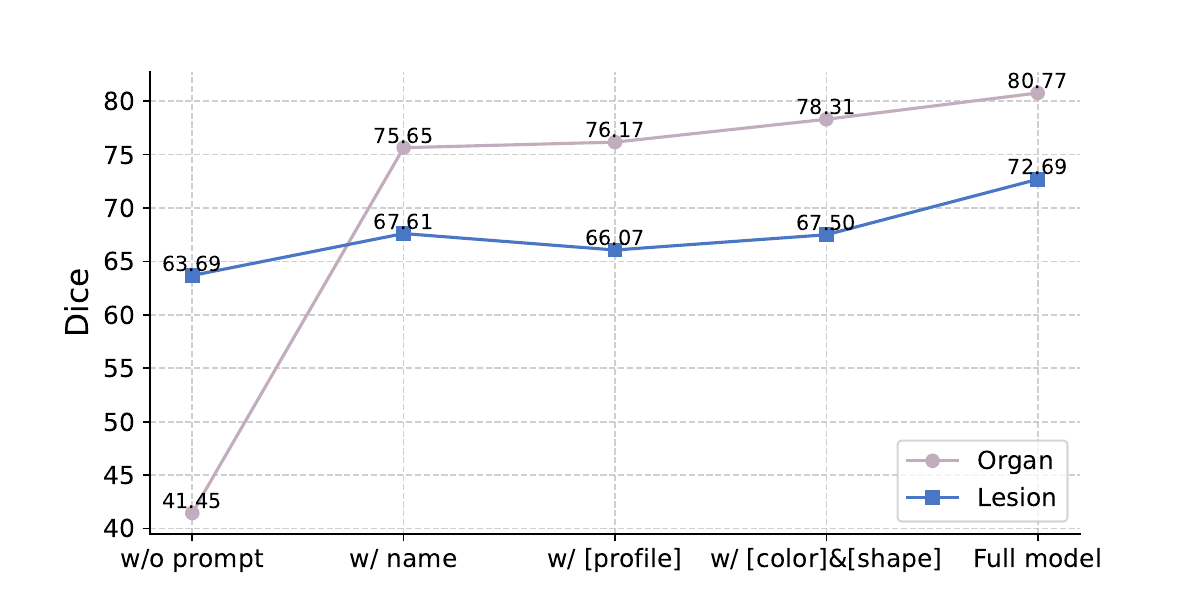}
    \caption{Ablation studies on different versions of medical text prompts.}
    \label{fig:line}
\end{figure}

\paragraph{Medical Text Prompts.} We utilize large language models to generate three attributes for each anatomical structure: [profile], [color], and [shape]. Among these, [profile] is a more abstract concept, whereas [color] and [shape] are more specific. These different attributes serve as varied prompt messages, resulting in distinct enhancements in segmentation performance, as shown in Figure~\ref{fig:line}.

We also conduct experiments with different prompt variations to evaluate their impact on segmentation performance.
For instance, simplified prompts with only class names result in Dice scores of 75.65\% for organs (-5.12\%) and 67.61\% for lesions (-5.08\%) compared to the full model. Examples of such simplified prompts include: ``an MRI of the myocardium'', ``a CT of the liver tumor'', ``an ultrasound image of the prostate''. The results demonstrate that detailed, descriptive prompts significantly enhance segmentation performance when compared to simplified ones.

\paragraph{Propagation Strategy.} To investigate the effects of box propagation, mask propagation, and query propagation, we conduct ablation experiments by removing the corresponding propagation methods, as demonstrated in Table~\ref{tab:propagation_main}. The absence of mask and query propagation results in decreases of 2.84 and 2.91 points in Dice score for organs. The results indicate that box propagation yields the smallest enhancements, with increases of 1.16 points for organs and 2.80 points for lesions in Dice scores. In contrast, mask and query propagation demonstrate a more significant impact, highlighting their critical roles in improving overall segmentation performance. This underscores the importance of designing appropriate propagation methods to optimize results in medical image sequence segmentation.

Table~\ref{tab:propagation2} analyzes the impact of different query selection strategies. The first row represents the case where no selection is performed. In the second row, the model selects the top-3 queries for Slice 2, and then the top-1 query for Slice 3. However, neither strategy outperforms the final configuration, indicating the effectiveness of retaining a single query across both Slice 2 and Slice 3.

\begin{table}[h]
\caption{Ablation studies on propagation.}
\centering
\renewcommand\arraystretch{1}
\setlength{\tabcolsep}{3pt}
\scalebox{1}{
\begin{tabular}{ccccc}
\toprule
\multirow{2}{*}{\makecell[c]{\textbf{Box} \\\textbf{propagation}}} & \multirow{2}{*}{\makecell[c]{\textbf{Mask} \\\textbf{propagation}}} & \multirow{2}{*}{\makecell[c]{\textbf{Query} \\\textbf{propagation}}} & \multirow{2}{*}{\textbf{Organ}} & \multirow{2}{*}{\textbf{Lesion}}   \\
& & & &  \\
\cmidrule(r){1-3} \cmidrule(){4-5}
\ding{55} & \ding{55} & \ding{55} & 74.53  & 63.97 \\
\ding{51} & \ding{51} & \ding{55} & 77.86  & 64.03 \\
\ding{51} & \ding{55} & \ding{51} & 77.93  & 67.10  \\
\ding{55} & \ding{51} & \ding{51} & 79.57  & 71.43  \\
  \ding{51} & \ding{51} & \ding{51} & \textbf{80.77} & \textbf{72.69} \\
\bottomrule
\end{tabular}}
\label{tab:propagation_main}
\end{table}

\begin{table}[h]
\caption{Analysis on query selection.}
\centering
\renewcommand\arraystretch{1.1}
\scalebox{1}{
\begin{tabular}{ccccc}
\toprule
\multicolumn{3}{c}{\textbf{Number of queries for}} & \multirow{2}{*}{\textbf{Organ}} & \multirow{2}{*}{\textbf{Lesion}} \\
\textbf{Slice 1}   & \textbf{Slice 2}  & \textbf{Slice 3}  &      &    \\
\cmidrule(r){1-3} \cmidrule(){4-5}
5 & 5 & 5  & 79.47   &  70.98      \\
5 & 3 & 1  & 78.47  &  71.67  \\
 5 & 1 & 1 & \textbf{80.77}   & \textbf{72.69}    \\
\bottomrule
\end{tabular}
}
\label{tab:propagation2}
\end{table}

\section{Conclusion}
In this paper, we introduce a new task, termed Referring Medical Image Sequence Segmentation, accompanied by a large and comprehensive benchmark. The benchmark includes 20 different anatomical structures across 4 modalities from various regions of the body. We present an innovative text-promptable approach that effectively leverages the inherent sequential relationships and textual cues within medical image sequences to segment referred objects, serving as a strong baseline for this task. By integrating both 2D and 3D medical images through a triple-propagation strategy, we demonstrate significant improvements across a broad spectrum of medical datasets, emphasizing the potential for rapid response in segmenting referred objects and enabling accurate diagnosis in clinical practice. Future work should delve deeper into optimizing prompts and exploring additional modalities to further enhance the efficacy of medical image analysis.


\bibliographystyle{ACM-Reference-Format}
\bibliography{sample-base}

\end{document}